\documentclass{article}

\usepackage{times}
\usepackage{graphicx} 
\usepackage{subfigure} 

\usepackage{natbib}

\usepackage{hyperref}

\usepackage{algorithm}
\usepackage{algorithmic}
\usepackage{amsmath}
\usepackage{amsfonts}
\usepackage{csquotes} 
\usepackage{hyperref}

\newcommand{\mb}[1]{\mathbf{\mathrm{#1}}}

\usepackage[accepted]{icml2017}

\icmltitlerunning{Neural Message Passing for Quantum Chemistry}

\begin{document} 

\twocolumn[
\icmltitle{Neural Message Passing for Quantum Chemistry}



\icmlsetsymbol{equal}{*}

\begin{icmlauthorlist}
\icmlauthor{Justin Gilmer}{brain}
\icmlauthor{Samuel S. Schoenholz}{brain}
\icmlauthor{Patrick F. Riley}{google}
\icmlauthor{Oriol Vinyals}{deepmind}
\icmlauthor{George E. Dahl}{brain}

\end{icmlauthorlist}

\icmlaffiliation{brain}{Google Brain}
\icmlaffiliation{google}{Google}
\icmlaffiliation{deepmind}{Google DeepMind}

\icmlcorrespondingauthor{Justin Gilmer}{gilmer@google.com}
\icmlcorrespondingauthor{George E. Dahl}{gdahl@google.com}

\icmlkeywords{deep learning, chemistry, DFT}

\vskip 0.3in
]



\printAffiliationsAndNotice{}  

\begin{abstract}

Supervised learning on molecules has incredible potential to be useful in chemistry, drug discovery, and materials science. Luckily, several promising and closely related neural network models invariant to molecular symmetries have already been described in the literature. These models learn a message passing algorithm and aggregation procedure to compute a function of their entire input graph. At this point, the next step is to find a particularly effective variant of this general approach and apply it to chemical prediction benchmarks until we either solve them or reach the limits of the approach. In this paper, we reformulate existing models into a single common framework we call Message Passing Neural Networks (MPNNs) and explore additional novel variations within this framework. Using MPNNs we demonstrate state of the art results on an important molecular property prediction benchmark; these results are strong enough that we believe future work should focus on datasets with larger molecules or more accurate ground truth labels.

\end{abstract}

\section{Introduction}

The past decade has seen remarkable success in the use of deep neural networks to understand and translate natural language~\cite{wu2016google}, generate and decode complex audio signals~\cite{hinton2012deep}, and infer features from real-world images and videos~\cite{krizhevsky2012imagenet}. Although chemists have applied machine learning to many problems over the years, predicting the properties of molecules and materials using machine learning (and especially deep learning) is still in its infancy. To date, most research applying machine learning to chemistry tasks ~\cite{bob,BAML,coloumb,rogers2010extended,montavon2012learning,behler2007,schoenholz2016}  has revolved around feature engineering. While neural networks have been applied in a variety of situations~\cite{merkwirth2005automatic, micheli2009neural, lusci2013deep, duvenaud2015}, they have yet to become widely adopted. This situation is reminiscent of the state of image models before the broad adoption of convolutional neural networks and is due, in part, to a dearth of empirical evidence that neural architectures with the appropriate inductive bias can be successful in this domain. 

\begin{figure}[t!]
\begin{center}
\label{fig:task}
\includegraphics[width=0.95\linewidth]{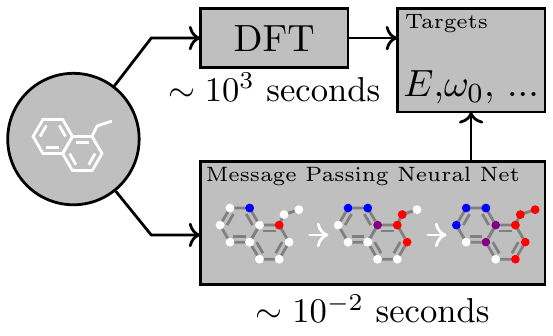}
\end{center}
\caption{A Message Passing Neural Network predicts quantum properties of an organic molecule by modeling a computationally expensive DFT calculation.}

\end{figure}

Recently, large scale quantum chemistry calculation and molecular dynamics simulations coupled with advances in high throughput experiments have begun to generate data at an unprecedented rate. Most classical techniques do not make effective use of the larger amounts of data that are now available. The time is ripe to apply more powerful and flexible machine learning methods to these problems, assuming we can find models with suitable inductive biases. The symmetries of atomic systems suggest neural networks that operate on graph structured data and are invariant to graph isomorphism might also be appropriate for molecules. Sufficiently successful models could someday help automate challenging chemical search problems in drug discovery or materials science.

In this paper, our goal is to demonstrate effective machine learning models for chemical prediction problems that are capable of learning their own features from molecular graphs directly and are invariant to graph isomorphism. To that end, we describe a general framework for supervised learning on graphs called Message Passing Neural Networks (MPNNs) that simply abstracts the commonalities between several of the most promising existing neural models for graph structured data, in order to make it easier to understand the relationships between them and come up with novel variations. Given how many researchers have published models that fit into the MPNN framework, we believe that the community should push this general approach as far as possible on practically important graph problems and only suggest new variations that are well motivated by applications, such as the application we consider here: predicting the quantum mechanical properties of small organic molecules (see task schematic in figure~\ref{fig:task}).

In general, the search for practically effective machine learning (ML) models in a given domain proceeds through a sequence of increasingly realistic and interesting benchmarks. Here we focus on the QM9 dataset as such a benchmark~\citep{ramakrishnan2014}. QM9 consists of 130k molecules with 13 properties for each molecule which are approximated by an expensive\footnote{By comparison, the inference time of the neural networks discussed in this work is 300k times faster.} quantum mechanical simulation method (DFT), to yield 13 corresponding regression tasks. These tasks are plausibly representative of many important chemical prediction problems and are (currently) difficult for many existing methods. Additionally, QM9 also includes complete spatial information for the single low energy conformation of the atoms in the molecule that was used in calculating the chemical properties. QM9 therefore lets us consider both the setting where the complete molecular geometry is known (atomic distances, bond angles, etc.) and the setting where we need to compute properties that might still be \emph{defined} in terms of the spatial positions of atoms, but where only the atom and bond information (i.e. graph) is available as input. In the latter case, the model must implicitly fit something about the computation used to determine a low energy 3D conformation and hopefully would still work on problems where it is not clear how to compute a reasonable 3D conformation.

When measuring the performance of our models on QM9, there are two important benchmark error levels. The first is the estimated average error of the DFT approximation to nature, which we refer to as ``DFT error.'' The  second, known as ``chemical accuracy,''  is a target error that has been established by the chemistry community. Estimates of DFT error and chemical accuracy are provided for each of the 13 targets in \citet{colab}. One important goal of this line of research is to produce a model which can achieve chemical accuracy with respect to the {\em true} targets as measured by an extremely precise experiment. The dataset containing the true targets on all 134k molecules does not currently exist. However, the ability to fit the DFT approximation to within chemical accuracy would be an encouraging step in this direction. For all 13 targets, achieving chemical accuracy is at least as hard as achieving DFT error. In the rest of this paper when we talk about chemical accuracy we generally mean with respect to our available ground truth labels.

%


In this paper, by exploring novel variations of models in the MPNN family, we are able to both achieve a new state of the art on the QM9 dataset and to predict the DFT calculation to within chemical accuracy on all but two targets. In particular, we provide the following key contributions:
\begin{itemize}
\item We develop an MPNN which achieves state of the art results on all 13 targets and predicts DFT to within chemical accuracy on 11 out of 13 targets.
\item We develop several different MPNNs which predict DFT to within chemical accuracy on 5 out of 13 targets while operating on the topology of the molecule alone (with no spatial information as input).
\item We develop a general method to train MPNNs with larger node representations without a corresponding increase in computation time or memory, yielding a substantial savings over previous MPNNs for high dimensional node representations.
\end{itemize}

We believe our work is an important step towards making well-designed MPNN’s the default for supervised learning on modestly sized molecules. In order for this to happen, researchers need to perform careful empirical studies to find the proper way to use these types of models and to make any necessary improvements to them, it is not sufficient for these models to have been described in the literature if there is only limited accompanying empirical work in the chemical domain. Indeed convolutional neural networks existed for decades before careful empirical work applying them to image classification~\citep{krizhevsky2012imagenet} helped them displace SVMs on top of hand-engineered features for a host of computer vision problems.

\section{Message Passing Neural Networks} \label{sec:mpnn}
There are at least eight notable examples of models from the literature that we can describe using our Message Passing Neural Networks (MPNN) framework. For simplicity we describe MPNNs which operate on undirected graphs $G$ with node features $x_v$ and edge features $e_{vw}$. It is trivial to extend the formalism to directed multigraphs. The forward pass has two phases, a message passing phase and a readout phase. The message passing phase runs for $T$ time steps and is defined in terms of message functions $M_t$ and vertex update functions $U_t$. During the message passing phase, hidden states $h_v^{t}$ at each node in the graph are updated based on messages $m_v^{t+1}$ according to

\begin{align} \label{eq:message}
   m_v^{t+1} & = \sum\limits_{w \in N(v)} M_t(h_v^t, h_w^t, e_{vw}) \\
\label{eq:update}
  h_v^{t+1} & = U_t( h_v^t, m_v^{t+1})
\end{align}



 where in the sum, $N(v)$ denotes the neighbors of $v$ in graph $G$. The readout phase computes a feature vector for the whole graph using some readout function $R$ according to
\begin{equation} \label{eq:output}
   \hat{y} = R(\{h_v^T \mid v \in G \}).
\end{equation}
The message functions $M_t$, vertex update functions $U_t$, and readout function $R$ are all learned differentiable functions. $R$ operates on the set of node states and must be invariant to permutations of the node states in order for the MPNN to be invariant to graph isomorphism. In what follows, we define previous models in the literature by specifying the message function $M_t$, vertex update function $U_t$, and readout function $R$ used. Note one could also learn edge features in an MPNN by introducing hidden states for all edges in the graph $h_{e_{vw}}^t$ and updating them analogously to equations \ref{eq:message} and \ref{eq:update}. Of the existing MPNNs, only \citet{kearnes2016molecular} has used this idea. 

\textbf{Convolutional Networks for Learning Molecular Fingerprints, \citet{duvenaud2015}}

The message function used is $M(h_v, h_w, e_{vw}) = (h_w, e_{vw})$ where $(.,.)$ denotes concatenation. The vertex update function used is $U_t(h_v^t, m_v^{t+1}) = \sigma(H_t^{\textrm{deg}(v)} m_v^{t+1})$, where $\sigma$ is the sigmoid function, deg$(v)$ is the degree of vertex $v$ and $H_t^N$ is a learned matrix for each time step $t$ and vertex degree $N$. $R$ has skip connections to all previous hidden states $h_v^t$ and is equal to $f \left( \sum\limits_{v, t} \textrm{softmax}( W_t h_v^t) \right)$, where $f$ is a neural network and $W_t$ are learned readout matrices, one for each time step $t$. This message passing scheme may be problematic since the resulting message vector is $m_v^{t+1} = \left( \sum h_w^t, \sum e_{vw} \right),$
which separately sums over connected nodes and connected edges. It follows that the message passing implemented in~\citet{duvenaud2015} is unable to identify correlations between edge states and node states.

\textbf{Gated Graph Neural Networks (GG-NN), \citet{yujia}}

The message function used is $M_t(h_v^t, h_w^t, e_{vw}) = A_{e_{vw}} h_w^t$, where $A_{e_{vw}}$ is a learned matrix, one for each edge label $e$ (the model assumes discrete edge types). The update function is $U_t = \textrm{GRU}(h_v^t, m_v^{t+1})$, where GRU is the Gated Recurrent Unit introduced in \citet{cho2014properties}. This work used weight tying, so the same update function is used at each time step $t$. Finally, 
\begin{equation} \label{eq:graph_level}
R =  \sum\limits_{v \in V} \sigma \left( i(h_v^{(T)}, h_v^0) \right) \odot \left( j(h_v^{(T)}) \right)
\end{equation}
  where $i$ and $j$ are neural networks, and $\odot$ denotes element-wise multiplication. 

\textbf{Interaction Networks, \citet{battaglia2016interaction}}

  This work considered both the case where there is a target at each node in the graph, and where there is a graph level target. It also considered the case where there are node level effects applied at each time step, in such a case the update function takes as input the concatenation $(h_v, x_v, m_v)$ where $x_v$ is an external vector representing some outside influence on the vertex $v$. 
  The message function $M(h_v, h_w, e_{vw})$ is a neural network which takes the concatenation $(h_v, h_w, e_{vw})$. The vertex update function $U(h_v,x_v, m_v)$ is a neural network which takes as input the concatenation $(h_v, x_v, m_v)$. Finally, in the case where there is a graph level output,  $R = f(\sum\limits_{v \in G} h_v^T)$ where $f$ is a neural network which takes the sum of the final hidden states $h_v^T$. Note the original work only defined the model for $T=1$.

\textbf{Molecular Graph Convolutions, \citet{kearnes2016molecular}}

This work deviates slightly from other MPNNs in that it introduces edge representations $e_{vw}^t$ which are updated during the message passing phase. The message function used for node messages is $M(h_v^t, h_w^t, e_{vw}^t) = e_{vw}^t$. The vertex update function is $U_t(h_v^t, m_v^{t+1}) = \alpha ( W_1 (\alpha(W_0 h_v^t), m_v^{t+1}) )$ where $(.,.)$ denotes concatenation, $\alpha$ is the ReLU activation and $W_1, W_0$ are learned weight matrices.  The edge state update is defined by $e_{vw}^{t+1} = U_t'(e_{vw}^t, h_v^t, h_w^t) =  \alpha ( W_4 
    (\alpha (W_2, e_{vw}^t) , \alpha (W_3 (h_v^t, h_w^t)) )  )$ where the $W_i$ are also learned weight matrices. 

\textbf{Deep Tensor Neural Networks, \citet{schutt2017quantum}}

    The message from $w$ to $v$ is computed by
    \[M_t = \textrm{tanh}\left( W^{fc}((W^{cf} h^t_w + b_1) \odot (W^{df} e_{vw} + b_2)) \right)\]
    where $W^{fc}$, $W^{cf}$, $W^{df}$ are matrices and $b_1$, $b_2$ are bias vectors. The update function used is $U_t(h_v^t, m_v^{t+1}) = h_v^t + m_v^{t+1}$. The readout function passes each node independently through a single hidden layer neural network and sums the outputs, in particular
    \[R = \sum\limits_{v} \textrm{NN}(h_v^{T}). \]

\textbf{Laplacian Based Methods, \citet{bruna2013spectral,defferrard2016convolutional,kipf2016}}

    These methods generalize the notion of the convolution operation typically applied to image datasets to an operation that operates on an arbitrary graph $G$ with a real valued adjacency matrix $A$. The operations defined in \citet{bruna2013spectral,defferrard2016convolutional} result in message functions of the form $M_t(h_v^t, h_w^t) = C^t_{vw} h_w^t$, where the matrices $C^t_{vw}$ are parameterized by the eigenvectors of the graph laplacian $L$, and the learned parameters of the model. The vertex update function used is $U_t(h_v^t, m_v^{t+1}) = \sigma(m_v^{t+1})$ where $\sigma$ is some pointwise non-linearity (such as ReLU).
    
    The \citet{kipf2016} model results in a message function $M_t(h_v^t, h_w^t) = c_{vw} h_w^t$ where $c_{vw} = \left(\textrm{deg}(v) \textrm{deg}(w)\right)^{-1/2} A_{vw}$. The vertex update function is $U_v^{t}(h_v^t, m_v^{t+1}) = \textrm{ReLU}(W^t m_v^{t+1})$. For the exact expressions for the $C^t_{vw}$ and the derivation of the reformulation of these models as MPNNs, see the supplementary material.

\subsection{ Moving Forward}
Given how many instances of MPNN’s have appeared in the literature, we should focus on pushing this general family as far as possible in a specific application of substantial practical importance. This way we can determine the most crucial implementation details and potentially reach the limits of these models to guide us towards future modeling improvements.

One downside of all of these approaches is computation time. Recent work has adapted the GG-NN architecture to larger graphs by passing messages on only subsets of the graph at each time step \citep{marino2016more}. In this work we also present a MPNN modification that can improve the computational costs.

\section{Related Work}

Although in principle quantum mechanics lets us compute the properties of molecules, the laws of physics lead to equations that are far too difficult to solve exactly. Therefore scientists have developed a hierarchy of approximations to quantum mechanics with varying tradeoffs of speed and accuracy, such as Density Functional Theory (DFT) with a variety of functionals~\citep{becke93, hohenberg64}, the GW approximation~\citep{hedin65}, and Quantum Monte-Carlo~\citep{ceperley86}. Despite being widely used, DFT is simultaneously still too slow to be applied to large systems (scaling as $\mathcal O(N_e^3)$ where $N_e$ is the number of electrons) and exhibits systematic as well as random errors relative to exact solutions to Schr\"odinger's equation. For example, to run the DFT calculation on a single 9 heavy atom molecule in QM9 takes around an hour on a single core of a Xeon E5-2660 (2.2 GHz) using  a version of Gaussian G09 (ES64L-G09RevD.01) ~\citep{bing2016pc}. For a 17 heavy atom molecule, computation time is up to 8 hours. Empirical potentials have been developed, such as the Stillinger-Weber potential~\citep{stillinger85}, that are fast and accurate but must be created from scratch, from first principles, for every new composition of atoms.

\citet{hu2003} used neural networks to approximate a particularly troublesome term in DFT called the exchange correlation potential to improve the accuracy of DFT. However, their method fails to improve upon the efficiency of DFT and relies on a large set of \textit{ad hoc} atomic descriptors.

Two more recent approaches by \citet{behler2007} and \citet{coloumb} attempt to approximate solutions to quantum mechanics directly without appealing to DFT. In the first case single-hidden-layer neural networks were used to approximate the energy and forces for configurations of a Silicon melt with the goal of speeding up molecular dynamics simulations. The second paper used Kernel Ridge Regression (KRR) to infer atomization energies over a wide range of molecules. In both cases hand engineered features were used (symmetry functions and the Coulomb matrix, respectively) that built physical symmetries into the input representation. Subsequent papers have replaced KRR by a neural network.

Both of these lines of research used hand engineered features that have intrinsic limitations. The work of~\citet{behler2007} used a representation that was manifestly invariant to graph isomorphism, but has difficulty when applied to systems with more than three species of atoms and fails to generalize to novel compositions. The representation used in~\citet{coloumb} is not invariant to graph isomorphism. Instead, this invariance must be learned by the downstream model through dataset augmentation.

In addition to the eight MPNN’s discussed in Section~\ref{sec:mpnn} there have been a number of other approaches to machine learning on graphical data which take advantage of the symmetries in a number of ways. One such family of approaches define a preprocessing step which constructs a canonical graph representation which can then be fed into into a standard classifier. Examples in this family include~\citet{niepert2016learning} and~\citet{coloumb}. Finally~\citet{scarselli2009graph} define a message passing process on graphs which is run until convergence, instead of for a finite number of time steps as in MPNNs.

\section{QM9 Dataset} 

To investigate the success of MPNNs on predicting chemical properties, we use the publicly available QM9 dataset \citep{ramakrishnan2014}. Molecules in the dataset consist of Hydrogen (H), Carbon (C), Oxygen (O), Nitrogen (N), and Flourine (F) atoms and contain up to 9 heavy (non Hydrogen) atoms. In all, this results in about 134k drug-like organic molecules that span a wide range of chemistry. For each molecule DFT is used to find a reasonable low energy  structure and hence atom ``positions'' are available. Additionally a wide range of interesting and fundamental chemical properties are computed. Given how fundamental some of the QM9 properties are, it is hard to believe success on more challenging chemical tasks will be possible if we can't make accurate statistical predictions for the properties computed in QM9.

We can group the different properties we try to predict into four broad categories. First, we have four properties related to how tightly bound together the atoms in a molecule are. These measure the energy required to break up the molecule at different temperatures and pressures. These include the atomization energy at $0K$, $U_0$ (eV), atomization energy at room temperature, $U$ (eV), enthalpy of atomization at room temperature, $H$ (eV), and free energy of atomization, $G$ (eV).

Next there are properties related to fundamental vibrations of the molecule, including the highest fundamental vibrational frequency $\omega_1$ ($cm^{-1}$) and the zero point vibrational energy (ZPVE) (eV). 

Additionally, there are a number of properties that concern the states of the electrons in the molecule. They include the energy of the electron in the highest occupied molecular orbital (HOMO) $\varepsilon_{\text{HOMO}}$ (eV), the energy of the lowest unoccupied molecular orbital (LUMO) $\varepsilon_{\text{LUMO}}$ (eV), and the electron energy gap ($\Delta\varepsilon$ (eV)). The electron energy gap is simply the difference $\varepsilon_{\text{HOMO}} - \varepsilon_{\text{LUMO}}$.  

Finally, there are several measures of the spatial distribution of electrons in the molecule. These include the electronic spatial extent $\langle R^2\rangle$ (Bohr$^2$), the norm of the dipole moment $\mu$ (Debye), and the norm of static polarizability $\alpha$ (Bohr$^3$). For a more detailed description of these properties, see the supplementary material.

\section{MPNN Variants} \label{sec:ggnn}

We began our exploration of MPNNs around the GG-NN model which we believe to be a strong baseline. We focused on trying different message functions, output functions, finding the appropriate input representation, and properly tuning hyperparameters.

For the rest of the paper we use $d$ to denote the dimension of the internal hidden representation of each node in the graph, and $n$ to denote the number of nodes in the graph. Our implementation of MPNNs in general operates on directed graphs with a separate message channel for incoming and outgoing edges, in which case the incoming message $m_v$ is the concatenation of $m_v^{\textrm{in}}$ and $m_v^{\textrm{out}}$, this was also used in \citet{yujia}. When we apply this to undirected chemical graphs we treat the graph as directed, where each original edge becomes both an incoming and outgoing edge with the same label. Note there is nothing special about the direction of the edge, it is only relevant for parameter tying. Treating undirected graphs as directed means that the size of the message channel is $2d$ instead of $d$. 

The input to our MPNN model is a set of feature vectors for the nodes of the graph, $x_v$, and an adjacency matrix $A$ with vector valued entries to indicate different bonds in the molecule as well as pairwise spatial distance between two atoms. We experimented as well with the message function used in the GG-NN family, which assumes discrete edge labels, in which case the matrix $A$ has entries in a discrete alphabet of size $k$. The initial hidden states $h_v^0$ are set to be the atom input feature vectors $x_v$ and are padded up to some larger dimension $d$. All of our experiments used weight tying at each time step $t$, and a GRU~\citep{cho2014properties} for the update function as in the GG-NN family. 

\subsection{Message Functions}

\textbf{Matrix Multiplication:}
	We started with the message function used in GG-NN which is defined by the equation $M(h_v, h_w, e_{vw}) = A_{e_{vw}} h_w$.
	
\textbf{Edge Network:}
	To allow vector valued edge features we propose the message function $M(h_v, h_w, e_{vw}) = A(e_{vw}) h_w$
 where $A(e_{vw})$ is a neural network which maps the edge vector $e_{vw}$ to a $d \times d$ matrix. 

\textbf{Pair Message:}
	   One property that the matrix multiplication rule has is that the message from node $w$ to node $v$ is a function only of the hidden state $h_w$ and the edge $e_{vw}$. In particular, it does not depend on the hidden state $h_v^t$. In theory, a network may be able to use the message channel more efficiently if the node messages are allowed to depend on both the source and destination node. Thus we also tried using a variant on the message function as described in \cite{battaglia2016interaction}. Here the message from $w$ to $v$ along edge $e$ is $m_{wv} = f\left(h_{w}^t, h_{v}^t, e_{vw}\right)$ where $f$ is a neural network.

When we apply the above message functions to directed graphs, there are two separate functions used, $M^{\textrm{in}}$ and an $M^{\textrm{out}}$. Which function is applied to a particular edge $e_{vw}$ depends on the direction of that edge.

\subsection{Virtual Graph Elements}
	We explored two different ways to change how the messages are passed throughout the model. The simplest modification involves adding a separate ``virtual'' edge type for pairs of nodes that are not connected. This can be implemented as a data preprocessing step and allows information to travel long distances during the propagation phase. 
	
  We also experimented with using a latent ``master'' node, which is connected to every input node in the graph with a special edge type. The master node serves as a global scratch space that each node both reads from and writes to in every step of message passing. We allow the master node to have a separate node dimension $d_{master}$, as well as separate weights for the internal update function (in our case a GRU). This allows information to travel long distances during the propagation phase. It also, in theory, allows additional model capacity (e.g. large values of $d_{master}$) without a substantial hit in performance, as the complexity of the master node model is $O(|E| d^2 + n d^2_{master})$. 

\subsection{Readout Functions}
   We experimented with two readout functions. First is the readout function used in GG-NN, which is defined by equation \ref{eq:graph_level}. Second is a set2set model from \citet{vinyals}. The set2set model is specifically designed to operate on sets and should have more expressive power than simply summing the final node states. This model first applies a linear projection to each tuple ($h_v^{T}, x_v$) and then takes as input the set of projected tuples $T = \{ (h_v^{T}, x_v)\}$. Then, after $M$ steps of computation, the set2set model produces a graph level embedding $q_t^*$ which is invariant to the order of the of the tuples $T$. We feed this embedding $q_t^*$ through a neural network to produce the output. 

\subsection{Multiple Towers} \label{subsec:towers}
   One issue with MPNN’s is scalability. In particular, a single step of the message passing phase for a dense graph requires $O( n^2 d^2)$ floating point multiplications. As $n$ or $d$ get large this can be computationally expensive. To address this issue we break the $d$ dimensional node embeddings $h_v^t$ into $k$ different $d/k$ dimensional embeddings $h_v^{t,k}$ and run a propagation step on each of the $k$ copies separately to get temporary embeddings $\{ \tilde{h}_v^{t+1,k}, v \in G\}$, using separate message and update functions for each copy. The $k$ temporary embeddings of each node are then mixed together according to the equation
 \begin{equation}
  \left(h_v^{t, 1}, h_v^{t, 2},\ldots, h_v^{t, k}\right) = g\left(\tilde{h}_v^{t, 1}, \tilde{h}_v^{t, 2}, \ldots , \tilde{h}_v^{t, k}\right)
\end{equation}
 where $g$ denotes a neural network and $(x, y, \ldots)$ denotes concatenation, with $g$ shared across all nodes in the graph. This mixing preserves the invariance to permutations of the nodes, while allowing the different copies of the graph to communicate with each other during the propagation phase. This can be advantageous in that it allows larger hidden states for the same number of parameters, which yields a computational speedup in practice. For example, when the message function is matrix multiplication (as in GG-NN) a propagation step of a single copy takes $O\left( n^2 (d/k)^2 \right)$ time, and there are $k$ copies, therefore the overall time complexity is $O \left(n^2 d^2 / k \right)$, with some additional overhead due to the mixing network. For $k=8$, $n=9$ and $d=200$ we see a factor of 2 speedup in inference time over a $k=1$, $n=9$, and $d=200$ architecture. This variation would be most useful for larger molecules, for instance molecules from GDB-17~\citep{ruddigkeit2012enumeration}.


\section{Input Representation} \label{subsec:input}
There are a number of features available for each atom in a molecule which capture both properties of the electrons in the atom as well as the bonds that the atom participates in. For a list of all of the features see table \ref{tb:features}. We experimented with making the hydrogen atoms explicit nodes in the graph (as opposed to simply including the count as a node feature), in which case graphs have up to 29 nodes. Note that having larger graphs significantly slows training time, in this case by a factor of roughly 10. For the adjacency matrix there are three edge representations used depending on the model. 

\begin{table}[t]
\centering
\caption{Atom Features}
\label{tb:features}
\begin{tabular}{lll}
\hline
\abovespace\belowspace
Feature             & Description                                      &  \\
\hline
\abovespace

Atom type           & H, C, N, O, F (one-hot)                  &  \\
Atomic number       & Number of protons (integer)                      &  \\
Acceptor            & Accepts electrons (binary)        &  \\
Donor               & Donates electrons (binary)        &  \\
Aromatic            & In an aromatic system (binary) &  \\
Hybridization       & sp, sp2, sp3 (one-hot or null)                   &  \\
Number of Hydrogens & (integer)        & 
\end{tabular}

\end{table}

\textbf{Chemical Graph:}
In the abscence of distance information, adjacency matrix entries are discrete bond types: single, double, triple, or aromatic.

\textbf{Distance bins:}
The matrix multiply message function assumes discrete edge types, so to include distance information we bin bond distances into 10 bins, the bins are obtained by uniformly partitioning the interval $[2,6]$ into 8 bins, followed by adding a bin $[0,2]$ and $[6,\infty]$. These bins were hand chosen by looking at a histogram of all distances. The adjacency matrix then has entries in an alphabet of size 14, indicating bond type for bonded atoms and distance bin for atoms that are not bonded. We found the distance for bonded atoms to be almost completely determined by bond type.

\textbf{Raw distance feature:}
When using a message function which operates on vector valued edges, the entries of the adjacency matrix are then 5 dimensional, where the first dimension indicates the euclidean distance between the pair of atoms, and the remaining four are a one-hot encoding of the bond type. 

\section{Training}

Each model and target combination was trained using a uniform random hyper parameter search with 50 trials. $T$ was constrained to be in the range $3 \leq T \leq 8$ (in practice, any $T \geq 3$ works). The number of set2set computations $M$ was chosen from the range $1 \leq M \leq 12$. All models were trained using SGD with the ADAM optimizer (\citet{kingma2014adam}), with batch size 20 for 3 million steps (~540 epochs). The initial learning rate was chosen uniformly between $1e^{-5}$ and $5e^{-4}$. We used a linear learning rate decay that began between 10\% and 90\% of the way through training and the initial learning rate $l$ decayed to a final learning rate $l*F$, using a decay factor $F$ in the range $[.01, 1]$.

The QM-9 dataset has 130462 molecules in it. We randomly chose 10000 samples for validation, 10000 samples for testing, and used the rest for training. We use the validation set to do early stopping and model selection and we report scores on the test set. All targets were normalized to have mean 0 and variance 1. We minimize the mean squared error between the model output and the target, although we evaluate mean absolute error. 

\section{Results}

   In all of our tables we report the ratio of the mean absolute error (MAE) of our models with the provided estimate of chemical accuracy for that target. Thus any model with error ratio less than 1 has achieved chemical accuracy for that target. In the supplementary material we list the chemical accuracy estimates for each target, these are the same estimates that were given in \citet{colab}. In this way, the MAE of our models can be calculated as $(\textrm{Error Ratio}) \times (\textrm{Chemical Accuracy}) $. Note, unless otherwise indicated, all tables display result of models trained individually on each target (as opposed to training one model to predict all 13).

\begin{table*}[t]
\centering
\caption{Comparison of Previous Approaches (left) with MPNN baselines (middle) and our methods (right)}
\abovespace\belowspace
\label{tb:main}
\begin{tabular}{llllll|lll|ll}
\hline
Target & BAML & BOB  & CM   & ECFP4  & HDAD & GC & GG-NN & DTNN & enn-s2s & enn-s2s-ens5 \\
\hline
\abovespace
mu     & 4.34 & 4.23 & 4.49 & 4.82   & 3.34   & 0.70 & 1.22  & - & \textbf{0.30}    & 0.20            \\
alpha  & 3.01 & 2.98 & 4.33 & 34.54  & 1.75   & 2.27 & 1.55  & - & \textbf{0.92}    & 0.68            \\
HOMO   & 2.20 & 2.20 & 3.09 & 2.89   & 1.54   & 1.18 & 1.17  & - & \textbf{0.99}    & 0.74            \\
LUMO   & 2.76 & 2.74 & 4.26 & 3.10   & 1.96   & 1.10 & 1.08  & - & \textbf{0.87}    & 0.65            \\
gap    & 3.28 & 3.41 & 5.32 & 3.86   & 2.49   & 1.78 & 1.70  & - & \textbf{1.60}    & 1.23            \\
R2     & 3.25 & 0.80 & 2.83 & 90.68  & 1.35   & 4.73 & 3.99  & - & \textbf{0.15}    & 0.14            \\
ZPVE   & 3.31 & 3.40 & 4.80 & 241.58 & 1.91   & 9.75 & 2.52  & - & \textbf{1.27}    & 1.10            \\
U0     & 1.21 & 1.43 & 2.98 & 85.01  & 0.58   & 3.02 & 0.83  & - & \textbf{0.45}    & 0.33            \\
U      & 1.22 & 1.44 & 2.99 & 85.59  & 0.59   & 3.16 & 0.86  & - & \textbf{0.45}    & 0.34            \\
H      & 1.22 & 1.44 & 2.99 & 86.21  & 0.59   & 3.19 & 0.81  & - & \textbf{0.39}    & 0.30            \\
G      & 1.20 & 1.42 & 2.97 & 78.36  & 0.59   & 2.95 & 0.78  & .84\footnotemark & \textbf{0.44}    & 0.34            \\
Cv     & 1.64 & 1.83 & 2.36 & 30.29  & 0.88   & 1.45 & 1.19  & - & \textbf{0.80}    & 0.62            \\
Omega  & 0.27 & 0.35 & 1.32 & 1.47   & 0.34   & 0.32 & 0.53  & - & \textbf{0.19}    & 0.15            \\
\hline
Average &	2.17 &	2.08 &	3.37 &	53.97 &	1.35 &		2.59 &	1.36 & - &	\textbf{0.68} &	0.52  \\
\end{tabular}
\end{table*}

\begin{table}[t]
\centering
\caption{Models Trained Without Spatial Information}
\label{tb:spatial}
\begin{tabular}{lll}
\hline
Model           & Average Error Ratio &  \\
\hline
GG-NN        & 3.47                &  \\
GG-NN + Virtual Edge        & 2.90                &  \\
GG-NN + Master Node & 2.62                &  \\
GG-NN + set2set    & \textbf{2.57}                & 
\end{tabular}

\end{table}

\begin{table}[t]
\centering
\caption{Towers vs Vanilla GG-NN (no explicit hydrogen)}
\label{tb:towers}
\begin{tabular}{lll}
\hline
Model          & Average Error Ratio &  \\
\hline
GG-NN + joint training    & 1.92                &  \\
towers8 + joint training & \textbf{1.75}               &  \\
\hline
GG-NN + individual training      & 1.53                &  \\
towers8 + individual training   & \textbf{1.37}             & 
\end{tabular}

\end{table}

We performed numerous experiments in order to find the best possible MPNN on this dataset as well as the proper input representation. In our experiments, we found that including the complete edge feature vector (bond type, spatial distance) and treating hydrogen atoms as explicit nodes in the graph to be very important for a number of targets. We also found that training one model per target consistently outperformed jointly training on all 13 targets. In some cases the improvement was up to 40\%. Our best MPNN variant used the edge network message function, set2set output, and operated on graphs with explicit hydrogens. We were able to further improve performance on the test set by ensembling the predictions of the five models with lowest validation error. 

In table \ref{tb:main} we compare the performance of our best MPNN variant (denoted with \textbf{enn-s2s}) and the corresponding ensemble (denoted with \textbf{enn-s2s-ens5}) with the previous state of the art on this dataset as reported in \citet{colab}. For clarity the error ratios of the best non-ensemble models are shown in bold. This previous work performed a comparison study of several existing ML models for QM9 and we have taken care to use the same train, validation, and test split. These baselines include 5 different hand engineered molecular representations, which then get fed through a standard, off-the-shelf classifier. These input representations include the Coulomb Matrix (\textbf{CM}, \citet{coloumb}), Bag of Bonds (\textbf{BoB}, \citet{bob}), Bonds Angles, Machine Learning (\textbf{BAML}, \citet{BAML}), Extended Connectivity Fingerprints (\textbf{ECPF4}, \citet{rogers2010extended}), and ``Projected Histograms'' (\textbf{HDAD}, \citet{colab}) representations. In addition to these hand engineered features we include two existing baseline MPNNs, the Molecular Graph Convolutions model (\textbf{GC}) from~\citet{kearnes2016molecular}, and the original GG-NN model~\citet{yujia} trained with distance bins. Overall, our new MPNN achieves chemical accuracy on 11 out of 13 targets and state of the art on all 13 targets.

\textbf{Training Without Spatial Information:} We also experimented in the setting where spatial information is not included in the input. In general, we find that augmenting the MPNN with some means of capturing long range interactions between nodes in the graph greatly improves performance in this setting. To demonstrate this we performed 4 experiments, one where we train the GG-NN model on the sparse graph, one where we add virtual edges, one where we add a master node, and one where we change the graph level output to a set2set output. The error ratios averaged across the 13 targets are shown in table~\ref{tb:spatial}. Overall, these three modifications help on all 13 targets, and the Set2Set output achieves chemical accuracy on 5 out of 13 targets. For more details, consult the supplementary material. The experiments shown tables~\ref{tb:spatial} and~\ref{tb:towers} were run with a partial charge feature as a node input. This feature is an output of the DFT calculation and thus could not be used in an applied setting. The state of art numbers we report in table~\ref{tb:main} do not use this feature.

\textbf{Towers:} Our original intent in developing the towers variant was to improve training time, as well as to allow the model to be trained on larger graphs. However, we also found some evidence that the multi-tower structure improves generalization performance. In table~\ref{tb:towers} we compare GG-NN + towers + set2set output vs a baseline GG-NN + set2set output when distance bins are used. We do this comparison in both the joint training regime and when training one model per target. 
The towers model outperforms the baseline model on 12 out of 13 targets in both individual and joint target training. We believe the benefit of towers is that it resembles training an ensemble of models. Unfortunately, our attempts so far at combining the towers and edge network message function have failed to further improve performance, possibly because the combination makes training more difficult. Further training details, and error ratios on all targets can be found in the supplementary material. 
\footnotetext{As reported in \citet{schutt2017quantum}. The model was trained on a different train/test split with 100k training samples vs 110k used in our experiments.}

\textbf{Additional Experiments:} In preliminary experiments, we tried disabling weight tying across different time steps. However, we found that the most effective way to increase performance was to tie the weights and use a larger hidden dimension $d$. We also early on found the pair message function to perform worse than the edge network function. This included a toy pathfinding problem which was originally designed to benefit from using pair messages. Also, when trained jointly on the 13 targets the edge network function outperforms pair message on 11 out of 13 targets, and has an average error ratio of 1.53 compared to 3.98 for pair message. Given the difficulties with training this function we did not pursue it further. For performance on smaller sized training sets, consult the supplementary material. 

\section{Conclusions and Future Work}

Our results show that MPNN’s with the appropriate message, update, and output functions have a useful inductive bias for predicting molecular properties, outperforming several strong baselines and eliminating the need for complicated feature engineering. Moreover, our results also reveal the importance of allowing long range interactions between nodes in the graph with either the master node or the set2set output. The towers variation makes these models more scalable, but additional improvements will be needed to scale to much larger graphs. 

An important future direction is to design MPNNs that can generalize effectively to larger graphs than those appearing in the training set or at least work with benchmarks designed to expose issues with generalization across graph sizes. Generalizing to larger molecule sizes seems particularly challenging when using spatial information. First of all, the pairwise distance distribution depends heavily on the number of atoms. Second, our most successful ways of using spatial information create a fully connected graph where the number of incoming messages also depends on the number of nodes. To address the second issue, we believe that adding an attention mechanism over the incoming message vectors could be an interesting direction to explore.

\section*{Acknowledgements} 

We would like to thank Lukasz Kaiser, Geoffrey Irving, Alex Graves, and Yujia Li for helpful discussions. Thank you to Adrian Roitberg for pointing out an issue with the use of partial charges in an earlier version of this paper.

\bibliography{mpnn}

\begin{thebibliography}{37}
\providecommand{\natexlab}[1]{#1}
\providecommand{\url}[1]{\texttt{#1}}
\expandafter\ifx\csname urlstyle\endcsname\relax
  \providecommand{\doi}[1]{doi: #1}\else
  \providecommand{\doi}{doi: \begingroup \urlstyle{rm}\Url}\fi

\bibitem[Battaglia et~al.(2016)Battaglia, Pascanu, Lai, Rezende, and
  Kavukcuoglu]{battaglia2016interaction}
Battaglia, Peter, Pascanu, Razvan, Lai, Matthew, Rezende, Danilo~Jimenez, and
  Kavukcuoglu, Koray.
\newblock Interaction networks for learning about objects, relations and
  physics.
\newblock In \emph{Advances in Neural Information Processing Systems}, pp.\
  4502--4510, 2016.

\bibitem[Becke(1993)]{becke93}
Becke, Axel~D.
\newblock Density-functional thermochemistry. iii. the role of exact exchange.
\newblock \emph{The Journal of Chemical Physics}, 98\penalty0 (7):\penalty0
  5648--5652, 1993.
\newblock \doi{10.1063/1.464913}.
\newblock URL \url{http://dx.doi.org/10.1063/1.464913}.

\bibitem[Behler \& Parrinello(2007)Behler and Parrinello]{behler2007}
Behler, J\"org and Parrinello, Michele.
\newblock Generalized neural-network representation of high-dimensional
  potential-energy surfaces.
\newblock \emph{Phys. Rev. Lett.}, 98:\penalty0 146401, Apr 2007.
\newblock \doi{10.1103/PhysRevLett.98.146401}.
\newblock URL \url{http://link.aps.org/doi/10.1103/PhysRevLett.98.146401}.

\bibitem[Bing et~al.(2017)Bing, von Lillenfeld, and Bakowies]{bing2016pc}
Bing, Huang, von Lillenfeld, O.~Anatole, and Bakowies, Dirk.
\newblock personal communication, 2017.

\bibitem[Bruna et~al.(2013)Bruna, Zaremba, Szlam, and LeCun]{bruna2013spectral}
Bruna, Joan, Zaremba, Wojciech, Szlam, Arthur, and LeCun, Yann.
\newblock Spectral networks and locally connected networks on graphs.
\newblock \emph{arXiv preprint arXiv:1312.6203}, 2013.

\bibitem[Ceperley \& Alder(1986)Ceperley and Alder]{ceperley86}
Ceperley, David and Alder, B.
\newblock Quantum monte carlo.
\newblock \emph{Science}, 231, 1986.

\bibitem[Cho et~al.(2014)Cho, Van~Merri{\"e}nboer, Bahdanau, and
  Bengio]{cho2014properties}
Cho, Kyunghyun, Van~Merri{\"e}nboer, Bart, Bahdanau, Dzmitry, and Bengio,
  Yoshua.
\newblock On the properties of neural machine translation: Encoder-decoder
  approaches.
\newblock \emph{arXiv preprint arXiv:1409.1259}, 2014.

\bibitem[Defferrard et~al.(2016)Defferrard, Bresson, and
  Vandergheynst]{defferrard2016convolutional}
Defferrard, Micha{\"e}l, Bresson, Xavier, and Vandergheynst, Pierre.
\newblock Convolutional neural networks on graphs with fast localized spectral
  filtering.
\newblock In \emph{Advances in Neural Information Processing Systems}, pp.\
  3837--3845, 2016.

\bibitem[Duvenaud et~al.(2015)Duvenaud, Maclaurin, Iparraguirre, Bombarell,
  Hirzel, Aspuru-Guzik, and Adams]{duvenaud2015}
Duvenaud, David~K, Maclaurin, Dougal, Iparraguirre, Jorge, Bombarell, Rafael,
  Hirzel, Timothy, Aspuru-Guzik, Al{\'a}n, and Adams, Ryan~P.
\newblock Convolutional networks on graphs for learning molecular fingerprints.
\newblock In \emph{Advances in neural information processing systems}, pp.\
  2224--2232, 2015.

\bibitem[Faber et~al.(2017)Faber, Hutchison, Huang, Gilmer, Schoenholz, Dahl,
  Vinyals, Kearnes, Riley, and von Lilienfeld]{colab}
Faber, Felix, Hutchison, Luke, Huang, Bing, Gilmer, Justin, Schoenholz,
  Samuel~S., Dahl, George~E., Vinyals, Oriol, Kearnes, Steven, Riley,
  Patrick~F., and von Lilienfeld, O.~Anatole.
\newblock Fast machine learning models of electronic and energetic properties
  consistently reach approximation errors better than dft accuracy.
\newblock \emph{https://arxiv.org/abs/1702.05532}, 2017.

\bibitem[Hansen et~al.(2015)Hansen, Biegler, Ramakrishnan, Pronobis, von
  Lilienfeld, Müller, and Tkatchenko]{bob}
Hansen, Katja, Biegler, Franziska, Ramakrishnan, Raghunathan, Pronobis, Wiktor,
  von Lilienfeld, O.~Anatole, Müller, Klaus-Robert, and Tkatchenko, Alexandre.
\newblock Machine learning predictions of molecular properties: Accurate
  many-body potentials and nonlocality in chemical space.
\newblock \emph{The journal of physical chemistry letters}, 6\penalty0
  (12):\penalty0 2326--2331, 2015.
\newblock \doi{10.1021/acs.jpclett.5b00831}.
\newblock URL \url{http://dx.doi.org/10.1021/acs.jpclett.5b00831}.

\bibitem[Hedin(1965)]{hedin65}
Hedin, Lars.
\newblock New method for calculating the one-particle green's function with
  application to the electron-gas problem.
\newblock \emph{Phys. Rev.}, 139:\penalty0 A796--A823, Aug 1965.
\newblock \doi{10.1103/PhysRev.139.A796}.
\newblock URL \url{http://link.aps.org/doi/10.1103/PhysRev.139.A796}.

\bibitem[Hinton et~al.(2012)Hinton, Deng, Yu, Dahl, Mohamed, Jaitly, Senior,
  Vanhoucke, Nguyen, Sainath, et~al.]{hinton2012deep}
Hinton, Geoffrey, Deng, Li, Yu, Dong, Dahl, George~E., Mohamed, Abdel-rahman,
  Jaitly, Navdeep, Senior, Andrew, Vanhoucke, Vincent, Nguyen, Patrick,
  Sainath, Tara~N, et~al.
\newblock Deep neural networks for acoustic modeling in speech recognition: The
  shared views of four research groups.
\newblock \emph{IEEE Signal Processing Magazine}, 29\penalty0 (6):\penalty0
  82--97, 2012.

\bibitem[Hohenberg \& Kohn(1964)Hohenberg and Kohn]{hohenberg64}
Hohenberg, P. and Kohn, W.
\newblock Inhomogeneous electron gas.
\newblock \emph{Phys. Rev.}, 136:\penalty0 B864--B871, Nov 1964.
\newblock \doi{10.1103/PhysRev.136.B864}.
\newblock URL \url{http://link.aps.org/doi/10.1103/PhysRev.136.B864}.

\bibitem[Hu et~al.(2003)Hu, Wang, Wong, and Chen]{hu2003}
Hu, LiHong, Wang, XiuJun, Wong, LaiHo, and Chen, GuanHua.
\newblock Combined first-principles calculation and neural-network correction
  approach for heat of formation.
\newblock \emph{The Journal of Chemical Physics}, 119\penalty0 (22):\penalty0
  11501--11507, 2003.

\bibitem[Huang \& von Lilienfeld(2016)Huang and von Lilienfeld]{BAML}
Huang, Bing and von Lilienfeld, O.~Anatole.
\newblock Communication: Understanding molecular representations in machine
  learning: The role of uniqueness and target similarity.
\newblock \emph{The Journal of Chemical Physics}, 145\penalty0 (16):\penalty0
  161102, 2016.
\newblock \doi{10.1063/1.4964627}.
\newblock URL \url{http://dx.doi.org/10.1063/1.4964627}.

\bibitem[Kearnes et~al.(2016)Kearnes, McCloskey, Berndl, Pande, and
  Riley]{kearnes2016molecular}
Kearnes, Steven, McCloskey, Kevin, Berndl, Marc, Pande, Vijay, and Riley,
  Patrick.
\newblock Molecular graph convolutions: Moving beyond fingerprints.
\newblock \emph{Journal of Computer-Aided Molecular Design}, 30\penalty0
  (8):\penalty0 595--608, 2016.

\bibitem[Kingma \& Ba(2014)Kingma and Ba]{kingma2014adam}
Kingma, Diederik and Ba, Jimmy.
\newblock Adam: A method for stochastic optimization.
\newblock \emph{arXiv preprint arXiv:1412.6980}, 2014.

\bibitem[{Kipf} \& {Welling}(2016){Kipf} and {Welling}]{kipf2016}
{Kipf}, T.~N. and {Welling}, M.
\newblock {Semi-Supervised Classification with Graph Convolutional Networks}.
\newblock \emph{ArXiv e-prints}, September 2016.

\bibitem[Krizhevsky et~al.(2012)Krizhevsky, Sutskever, and
  Hinton]{krizhevsky2012imagenet}
Krizhevsky, Alex, Sutskever, Ilya, and Hinton, Geoffrey~E.
\newblock Imagenet classification with deep convolutional neural networks.
\newblock In \emph{Advances in neural information processing systems}, pp.\
  1097--1105, 2012.

\bibitem[Li et~al.(2016)Li, Tarlow, Brockschmidt, and Zemel]{yujia}
Li, Yujia, Tarlow, Daniel, Brockschmidt, Marc, and Zemel, Richard.
\newblock Gated graph sequence neural networks.
\newblock \emph{ICLR}, 2016.

\bibitem[Lusci et~al.(2013)Lusci, Pollastri, and Baldi]{lusci2013deep}
Lusci, Alessandro, Pollastri, Gianluca, and Baldi, Pierre.
\newblock Deep architectures and deep learning in chemoinformatics: the
  prediction of aqueous solubility for drug-like molecules.
\newblock \emph{Journal of chemical information and modeling}, 53\penalty0
  (7):\penalty0 1563--1575, 2013.

\bibitem[Marino et~al.(2016)Marino, Salakhutdinov, and Gupta]{marino2016more}
Marino, Kenneth, Salakhutdinov, Ruslan, and Gupta, Abhinav.
\newblock The more you know: Using knowledge graphs for image classification.
\newblock \emph{arXiv preprint arXiv:1612.04844}, 2016.

\bibitem[Merkwirth \& Lengauer(2005)Merkwirth and
  Lengauer]{merkwirth2005automatic}
Merkwirth, Christian and Lengauer, Thomas.
\newblock Automatic generation of complementary descriptors with molecular
  graph networks.
\newblock \emph{Journal of chemical information and modeling}, 45\penalty0
  (5):\penalty0 1159--1168, 2005.

\bibitem[Micheli(2009)]{micheli2009neural}
Micheli, Alessio.
\newblock Neural network for graphs: A contextual constructive approach.
\newblock \emph{IEEE Transactions on Neural Networks}, 20\penalty0
  (3):\penalty0 498--511, 2009.

\bibitem[Montavon et~al.(2012)Montavon, Hansen, Fazli, Rupp, Biegler, Ziehe,
  Tkatchenko, von Lilienfeld, and M{\"u}ller]{montavon2012learning}
Montavon, Gr{\'e}goire, Hansen, Katja, Fazli, Siamac, Rupp, Matthias, Biegler,
  Franziska, Ziehe, Andreas, Tkatchenko, Alexandre, von Lilienfeld, O.~Anatole,
  and M{\"u}ller, Klaus-Robert.
\newblock Learning invariant representations of molecules for atomization
  energy prediction.
\newblock In \emph{Advances in Neural Information Processing Systems}, pp.\
  440--448, 2012.

\bibitem[Niepert et~al.(2016)Niepert, Ahmed, and Kutzkov]{niepert2016learning}
Niepert, Mathias, Ahmed, Mohamed, and Kutzkov, Konstantin.
\newblock Learning convolutional neural networks for graphs.
\newblock In \emph{Proceedings of the 33rd annual international conference on
  machine learning. ACM}, 2016.

\bibitem[Ramakrishnan et~al.(2014)Ramakrishnan, Dral, Rupp, and
  Von~Lilienfeld]{ramakrishnan2014}
Ramakrishnan, Raghunathan, Dral, Pavlo~O, Rupp, Matthias, and Von~Lilienfeld,
  O~Anatole.
\newblock Quantum chemistry structures and properties of 134 kilo molecules.
\newblock \emph{Scientific data}, 1, 2014.

\bibitem[Rogers \& Hahn(2010)Rogers and Hahn]{rogers2010extended}
Rogers, David and Hahn, Mathew.
\newblock Extended-connectivity fingerprints.
\newblock \emph{Journal of chemical information and modeling}, 50\penalty0
  (5):\penalty0 742--754, 2010.

\bibitem[Ruddigkeit et~al.(2012)Ruddigkeit, Van~Deursen, Blum, and
  Reymond]{ruddigkeit2012enumeration}
Ruddigkeit, Lars, Van~Deursen, Ruud, Blum, Lorenz~C, and Reymond, Jean-Louis.
\newblock Enumeration of 166 billion organic small molecules in the chemical
  universe database gdb-17.
\newblock \emph{Journal of chemical information and modeling}, 52\penalty0
  (11):\penalty0 2864--2875, 2012.

\bibitem[Rupp et~al.(2012)Rupp, Tkatchenko, and von Lilienfeld]{coloumb}
Rupp, Matthias, Tkatchenko, Alexandre haand~M\"uller, Klaus-Robert, and von
  Lilienfeld, O.~Anatole.
\newblock Fast and accurate modeling of molecular atomization energies with
  machine learning.
\newblock \emph{Physical review letters}, 108\penalty0 (5):\penalty0 058301,
  Jan 2012.
\newblock URL \url{http://dx.doi.org/10.1103/PhysRevLett.108.058301}.

\bibitem[Scarselli et~al.(2009)Scarselli, Gori, Tsoi, Hagenbuchner, and
  Monfardini]{scarselli2009graph}
Scarselli, Franco, Gori, Marco, Tsoi, Ah~Chung, Hagenbuchner, Markus, and
  Monfardini, Gabriele.
\newblock The graph neural network model.
\newblock \emph{IEEE Transactions on Neural Networks}, 20\penalty0
  (1):\penalty0 61--80, 2009.

\bibitem[Schoenholz et~al.(2016)Schoenholz, Cubuk, Sussman, Kaxiras, and
  Liu]{schoenholz2016}
Schoenholz, Samuel~S., Cubuk, Ekin~D., Sussman, Daniel~M, Kaxiras, Efthimios,
  and Liu, Andrea~J.
\newblock A structural approach to relaxation in glassy liquids.
\newblock \emph{Nature Physics}, 2016.

\bibitem[Sch{\"u}tt et~al.(2017)Sch{\"u}tt, Arbabzadah, Chmiela, M{\"u}ller,
  and Tkatchenko]{schutt2017quantum}
Sch{\"u}tt, Kristof~T, Arbabzadah, Farhad, Chmiela, Stefan, M{\"u}ller,
  Klaus~R, and Tkatchenko, Alexandre.
\newblock Quantum-chemical insights from deep tensor neural networks.
\newblock \emph{Nature Communications}, 8, 2017.

\bibitem[Stillinger \& Weber(1985)Stillinger and Weber]{stillinger85}
Stillinger, Frank~H. and Weber, Thomas~A.
\newblock Computer simulation of local order in condensed phases of silicon.
\newblock \emph{Phys. Rev. B}, 31:\penalty0 5262--5271, Apr 1985.
\newblock \doi{10.1103/PhysRevB.31.5262}.
\newblock URL \url{http://link.aps.org/doi/10.1103/PhysRevB.31.5262}.

\bibitem[Vinyals et~al.(2015)Vinyals, Bengio, and Kudlur]{vinyals}
Vinyals, Oriol, Bengio, Samy, and Kudlur, Manjunath.
\newblock Order matters: Sequence to sequence for sets.
\newblock \emph{arXiv preprint arXiv:1511.06391}, 2015.

\bibitem[Wu et~al.(2016)Wu, Schuster, Chen, Le, Norouzi, Macherey, Krikun, Cao,
  Gao, Macherey, et~al.]{wu2016google}
Wu, Yonghui, Schuster, Mike, Chen, Zhifeng, Le, Quoc~V., Norouzi, Mohammad,
  Macherey, Wolfgang, Krikun, Maxim, Cao, Yuan, Gao, Qin, Macherey, Klaus,
  et~al.
\newblock Google's neural machine translation system: Bridging the gap between
  human and machine translation.
\newblock \emph{arXiv preprint arXiv:1609.08144}, 2016.

\end{thebibliography}
\bibliographystyle{icml2017}

\section{Appendix}
\subsection{Interpretation of Laplacian based models as MPNNs}
    Another family of models defined in  \citet{defferrard2016convolutional}, \citet{bruna2013spectral}, \citet{kipf2016} can be interpreted as MPNNs. These models generalize the notion of convolutions a general graph $G$ with $N$ nodes. In the language of MPNNs, these models tend to have very simple message functions and are typically applied to much larger graphs such as social network data.  We closely follow the notation defined in \citet{bruna2013spectral} equation (3.2). The model discussed in \citet{defferrard2016convolutional} (equation 5) and \citet{kipf2016} can be viewed as special cases.  Given an adjacency matrix $W \in \mathbb{R}^{N \times N}$ we define the graph Laplacian to be $L = I_n - D^{-1/2} W D^{-1/2}$ where $D$ is the diagonal degree matrix with $D_{ii} = \textrm{deg}(v_i)$. Let $V$ denote the eigenvectors of $L$, ordered by eigenvalue. Let $\sigma$ be a real valued nonlinearity (such as ReLU). We now define an operation which transforms an input vector $x$ of size $N \times d_1$ to a vector $y$ of size $N \times d_2$ (the full model can defined as stacking these operations). 
    
    \begin{equation} \label{eq:eq1}
        y_j = \sigma \left( \sum\limits_{i=1}^{d_1} V F_{i,j} V^T x_i \right) \quad (j = 1 \ldots d_2)
    \end{equation}
    Here $y_j$ and $x_i$ are all $N$ dimensional vectors corresponding to a scalar feature at each node. The matrices $F_{i,j}$ are all diagonal $N \times N$ matrices and contain all of the learned parameters in the layer. We now expand equation \ref{eq:eq1} in terms of the full $N \times d_1$ vector $x$ and $N \times d_2$ vector $y$, using $v$ and $w$ to index nodes in the graph $G$ and $i$, $j$ to index the dimensions of the node states. In this way $x_{w,i}$ denotes the $i$'th dimension of node $w$, and $y_{v,j}$ denotes the $j$'th dimension of node $v$, furthermore we use $x_w$ to denote the $d_1$ dimensional vector for node state $w$, and $y_v$ to denote the $d_2$ dimensional vector for node $v$. Define the rank 4 tensor $\tilde{L}$ of dimension $N \times N \times d_1 \times d_2$ where $\tilde{L}_{v,w,i,j} = (V F_{i,j} V^T)_{v,w}$. We will use $\tilde{L}_{i,j}$ to denote the $N \times N$ dimensional matrix where  $(\tilde{L}_{i,j})_{v,w} = \tilde{L}_{v,w,i,j}$ and $\tilde{L}_{v,w}$ to denote the $d_1 \times d_2$ dimensional matrix where $(\tilde{L}_{v,w})_{i,j} = \tilde{L}_{v,w,i,j}$. Writing equation~\ref{eq:eq1} in this notation we have
    
    \begin{align*} 
        y_j & = \sigma \left( \sum\limits_{i=1}^{d_1} \tilde{L}_{i,j} x_i \right) \\
        y_{v,j} & = \sigma \left( \sum\limits_{i=1,w=1}^{d_1,N} \tilde{L}_{v,w,i,j} x_{w,i}\right) \\
        y_{v} & = \sigma \left( \sum\limits_{w=1}^{N} \tilde{L}_{v,w} x_{w} \right). \\
    \end{align*}
    Relabelling $y_v$ as $h_v^{t+1}$ and $x_w$ as $h_w^t$ this last line can be interpreted as the message passing update in an MPNN where $M(h_v^t, h_w^t) = \tilde{L}_{v,w} h_w^t$ and $U(h^t_v, m^{t+1}_v) = \sigma(m^{t+1}_v)$. 
    
    \subsubsection{The special case of Kipf and Welling (2016)}
    
    Motivated as a first order approximation of the graph laplacian methods, \citet{kipf2016} propose the following layer-wise propagation rule:
    \begin{equation}
        H^{l+1} = \sigma \left(\tilde{D}^{-1/2} \tilde{A} \tilde{D}^{-1/2} H^l W^l\right)
    \end{equation}
    Here $\tilde{A} = A + I_N$ where $A$ is the real valued adjacency matrix for an undirected graph $G$. Adding the identity matrix $I_N$ corresponds to adding self loops to the graph. Also $\tilde{D}_{ii} = \sum_j \tilde{A}_{ij}$ denotes the degree matrix for the graph with self loops, $W^{l} \in \mathbb{R}^{D \times D}$ is a layer-specific trainable weight matrix, and $\sigma(\cdot)$ denotes a real valued nonlinearity. Each $H^{l}$ is a $\mathbb{R}^{N \times D}$ dimensional matrix indicating the $D$ dimensional node states for the $N$ nodes in the graph. 
    
    In what follows, given a matrix $M$ we use $M_{(v)}$ to denote the row in $M$ indexed by $v$ ($v$ will always correspond to a node in the graph $G$). Let $L = \tilde{D}^{-1/2} \tilde{A} \tilde{D}^{-1/2}$. To get the updated node state for node $v$ we have 
    \begin{align*}
        H^{l+1}_{(v)} & = \sigma\left( L_{(v)} H^l W^l\right) \\
        & = \sigma\left( \sum\limits_{w} L_{vw} H^{l}_{(w)} W^{l} \right) \\
    \end{align*}
    Relabelling the row vector $H^{l+1}_{(v)}$ as an $N$ dimensional column vector $h_v^{t+1}$ the above equation is equivalent to
    \begin{equation}
        h^{t+1}_v = \sigma\left((W^{l})^T \sum\limits_{w} L_{vw} h_w^{t} \right)
    \end{equation}
    which is equivalent to a message function 
    \[ M_t(h^t_v, h^t_w) = L_{vw} h_w^{t} = \tilde{A}_{vw} (\textrm{deg}(v) \textrm{deg}(w))^{-1/2} h^t_w, \]
    and update function 
    \[U_t(h^t_v, m^{t+1}_v) = \sigma((W^{t})^T m^{t+1}).\]
    Note that the $L_{vw}$ are all scalar valued, so this model corresponds to taking a certain weighted average of neighboring nodes at each time step. 
    
\subsection{A More Detailed Description of the Quantum Properties}

First there the four atomization energies. 

\begin{itemize}
    \item Atomization energy at $0K$ $U_0$ (eV): This is the energy required to break up the molecule into all of its constituent atoms if the molecule is at absolute zero. This calculation assumes that the molecules are held at fixed volume.
    \item Atomization energy at room temperature $U$ (eV): Like $U_0$, this is the energy required to break up the molecule if it is at room temperature. 
    \item Enthalpy of atomization at room temperature $H$ (eV): The enthalpy of atomization is similar in spirit to the energy of atomization, $U$. However, unlike the energy this calculation assumes that the constituent molecules are held at fixed pressure.
    \item Free energy of atomization $G$ (eV): Once again this is similar to $U$ and $H$, but assumes that the system is held at fixed temperature and pressure during the dissociation. 
\end{itemize}
Next there are properties related to fundamental vibrations of the molecule:
\begin{itemize}
    \item Highest fundamental vibrational frequency $\omega_1$ ($cm^{-1}$): Every molecule has fundamental vibrational modes that it can naturally oscillate at. $\omega_1$ is the mode that requires the most energy.
    \item Zero Point Vibrational Energy (ZPVE) (eV): Even at zero temperature quantum mechanical uncertainty implies that atoms vibrate. This is known as the zero point vibrational energy and can be calculated once the allowed vibrational modes of a molecule are known.
\end{itemize}
Additionally, there are a number of properties that concern the states of the electrons in the molecule:
\begin{itemize}
    \item Highest Occupied Molecular Orbital (HOMO) $\varepsilon_{\text{HOMO}}$ (eV): Quantum mechanics dictates that the allowed states that electrons can occupy in a molecule are discrete. The Pauli exclusion principle states that no two electrons may occupy the same state. At zero temperature, therefore, electrons stack in states from lowest energy to highest energy. HOMO is the energy of the highest occupied electronic state.
    \item Lowest Unoccupied Molecular Orbital (LUMO) $\varepsilon_{\text{LUMO}}$ (eV): Like HOMO, LUMO is the lowest energy electronic state that is unoccupied.
    \item Electron energy gap $\Delta\varepsilon$ (eV): This is the difference in energy between LUMO and HOMO. It is the lowest energy transition that can occur when an electron is excited from an occupied state to an unoccupied state. $\Delta\varepsilon$ also dictates the longest wavelength of light that the molecule can absorb.
\end{itemize}
Finally, there are several measures of the spatial distribution of electrons in the molecule:
\begin{itemize}
    \item Electronic Spatial Extent $\langle R^2\rangle$ (Bohr$^2$): The electronic spatial extent is the second moment of the charge distribution, $\rho(r)$, or in other words $\langle R^2\rangle = \int dr r^2\rho(r)$.
    \item Norm of the dipole moment $\mu$ (Debye): The dipole moment, $p(r) = \int dr'p(r')(r-r')$, approximates the electric field far from a molecule. The norm of the dipole moment is related to how anisotropically the charge is distributed (and hence the strength of the field far from the molecule). This degree of anisotropy is in turn related to a number of material properties (for example hydrogen bonding in water causes the dipole moment to be large which has a large effect on dynamics and surface tension).
    \item Norm of the static polarizability $\alpha$ (Bohr$^3$): $\alpha$ measures the extent to which a molecule can spontaneously incur a dipole moment in response to an external field. This is in turn related to the degree to which i.e. Van der Waals interactions play a role in the dynamics of the medium. 
\end{itemize}

\subsection{Chemical Accuracy and DFT Error}
    In Table~\ref{tb:chem_MAE} we list the mean absolute error numbers for chemical accuracy. These are the numbers used to compute the error ratios of all models in the tables. The mean absolute errors of our models can thus be calculated as $(\textrm{Error Ratio}) \times (\textrm{Chemical Accuracy})$. We also include some estimates of the mean absolute error for the DFT calculation to the ground truth. Both the estimates of chemical accruacy and DFT error were provided in \citet{colab}.

\begin{table}[t]
\centering
\caption{Chemical Accuracy For Each Target}
\label{tb:chem_MAE}
\begin{tabular}{lllll}
\hline
Target & DFT Error  & Chemical Accuracy \\
\hline

mu     & .1 & .1         \\
alpha  & .4 & .1           \\
HOMO   & 2.0 & .043          \\
LUMO   & 2.6 & .043       \\
gap    & 1.2 & .043        \\
R2     & - & 1.2         \\
ZPVE   & .0097 & .0012        \\
U0     & .1 & .043         \\
U      & .1 & .043           \\
H      & .1 &.043          \\
G      & .1 & .043      \\
Cv     & .34 & .050        \\
Omega  & 28 & 10.0          \\
\end{tabular}
\end{table}

\subsection{Additional Results}

In Table~\ref{tb:learning_curve} we compare the performance of the best architecture (edge network + set2set output) on different sized training sets. It is surprising how data efficient this model is on some targets. For example, on both R2 and Omega our models are equal or better with 11k samples than the best baseline is with 110k samples.

\begin{table*}[t]
\centering
\caption{Results from training the edge network + set2set model on different sized training sets (N denotes the number of training samples)}
\label{tb:learning_curve}
\begin{tabular}{llllll}
\hline
Target & N=11k & N=35k & N=58k & N=82k & N=110k \\
\hline
mu     & 1.28    & 0.55    & 0.44    & 0.32    & 0.30     \\
alpha  & 2.76    & 1.59    & 1.26    & 1.09    & 0.92     \\
HOMO   & 2.33    & 1.50    & 1.34    & 1.19    & 0.99     \\
LUMO   & 2.18    & 1.47    & 1.19    & 1.10    & 0.87     \\
gap    & 3.53    & 2.34    & 2.07    & 1.84    & 1.60     \\
R2     & 0.28    & 0.22    & 0.21    & 0.21    & 0.15     \\
ZPVE   & 2.52    & 1.78    & 1.69    & 1.68    & 1.27     \\
U0     & 1.24    & 0.69    & 0.58    & 0.62    & 0.45     \\
U      & 1.05    & 0.69    & 0.60    & 0.52    & 0.45     \\
H      & 1.14    & 0.64    & 0.65    & 0.53    & 0.39     \\
G      & 1.23    & 0.62    & 0.64    & 0.49    & 0.44     \\
Cv     & 1.99    & 1.24    & 0.93    & 0.87    & 0.80     \\
Omega  & 0.28    & 0.25    & 0.24    & 0.15    & 0.19    
\end{tabular}
\end{table*}

In Table~\ref{tb:spatial2} we compare the performance of several models trained without spatial information. The left 4 columns show the results of 4 experiments, one where we train the GG-NN model on the sparse graph, one where we add virtual edges (\textbf{ve}), one where we add a master node (\textbf{mn}), and one where we change the graph level output to a set2set output (\textbf{s2s}). In general, we find that it's important to allow the model to capture long range interactions in these graphs. 

In Table~\ref{tb:towers2} we compare GG-NN + towers + set2set output (\textbf{tow8}) vs a baseline GG-NN + set2set output (\textbf{GG-NN}) when distance bins are used.  We do this comparison in both the joint training regime (\textbf{j}) and when training one model per target (\textbf{i}). For joint training of the baseline we used 100 trials with $d=200$ as well as 200 trials where $d$ was chosen randomly in the set $\{43, 73, 113, 153\}$, we report the minimum test error across all 300 trials. For individual training of the baseline we used 100 trials where $d$ was chosen uniformly in the range $[43,200]$. The towers model was always trained with $d = 200$ and $k=8$, with 100 tuning trials for joint training and 50 trials for individual training. The towers model outperforms the baseline model on 12 out of 13 targets in both individual and joint target training.

In Table~\ref{tb:pm2} right 2 columns compare the edge network (\textbf{enn}) with the pair message network (\textbf{pm}) in the joint training regime (\textbf{j}). The edge network consistently outperforms the pair message function across most targets. 

In Table~\ref{tb:input} we compare our MPNNs with different input featurizations. All models use the Set2Set output and GRU update functions. The no distance model uses the matrix multiply message function, the distance models use the edge neural network message function.

\begin{table}[t]
\centering
\caption{Comparison of models when distance information is excluded}
\label{tb:spatial2}
\begin{tabular}{lllll}
\hline
Target & GG-NN  & ve  & mn & s2s \\
\hline

mu     & 3.94     & \textbf{3.76}     & 4.02            & 3.81        \\
alpha  & 2.43     & 2.07     & \textbf{2.01}            & 2.04        \\
HOMO   & 1.80     & \textbf{1.60}     & 1.67            & 1.71        \\
LUMO   & 1.73     & 1.48     & 1.48            & \textbf{1.41}        \\
gap    & 2.48     & 2.33     & \textbf{2.23}            & 2.26        \\
R2     & 14.74    & 17.11    & \textbf{13.16}           & 13.67       \\
ZPVE   & 5.93     & 3.21     & 3.26            & \textbf{3.02}        \\
U0     & 1.98     & 0.89     & 0.90            & \textbf{0.72}        \\
U      & 2.48     & 0.93     & 0.99            & \textbf{0.79}        \\
H      & 2.19     & 0.86     & 0.95            & \textbf{0.80}        \\
G      & 2.13     & 0.85     & 1.02            & \textbf{0.74}        \\
Cv     & 1.96     & \textbf{1.61}     & 1.63            & 1.71        \\
Omega  & 1.28     & 1.05     & 0.78            & \textbf{0.78}        \\
\hline
Average & 3.47 & 2.90 & 2.62 & \textbf{2.57}
\end{tabular}
\end{table}

\begin{table}[t]
\centering
\caption{Towers vs Vanilla Model (no explicit hydrogen)}
\label{tb:towers2}
\begin{tabular}{l|ll|ll}
\hline
Target & GG-NN-j & tow8-j & GG-NN-i & tow8-i \\
\hline
mu & 2.73        & \textbf{2.47}           & \textbf{2.16}      & 2.23         \\
alpha & 1.66        & \textbf{1.50}           & 1.47      & \textbf{1.34}         \\
HOMO & 1.33        & \textbf{1.19}           & 1.27      & \textbf{1.20}         \\
LUMO & 1.28        & \textbf{1.12}           & 1.24      & \textbf{1.11}         \\
gap & 1.73        & \textbf{1.55}           & 1.78      & \textbf{1.68}         \\
R2 & \textbf{6.07}   & 6.16           & 4.78      & \textbf{4.11}         \\
ZPVE & 4.07        & \textbf{3.43}           & 2.70      & \textbf{2.54}         \\
U0 & 1.00        & \textbf{0.86}           & 0.71      & \textbf{0.55}         \\
U & 1.01        & \textbf{0.88}           & 0.65      & \textbf{0.52}         \\
H & 1.01        & \textbf{0.88}           & 0.68      & \textbf{0.50}         \\
G & 0.99        & \textbf{0.85}           & 0.66      & \textbf{0.50}         \\
Cv & 1.40        & \textbf{1.27}           & 1.27      & \textbf{1.09}         \\
Omega & 0.66        & \textbf{0.62}           & 0.57      & \textbf{0.50}         \\
\hline
Average & 1.92 & \textbf{1.75} & 1.53 & \textbf{1.37}
\end{tabular}
\end{table}

\begin{table}[t]
\centering
\caption{Pair Message vs Edge Network in joint training}
\label{tb:pm2}
\begin{tabular}{lll}
\hline
Target & pm-j & enn-j \\
\hline

mu & \textbf{2.10}     & 2.24      \\
alpha & 2.30     & \textbf{1.48}      \\
HOMO & \textbf{1.20}     & 1.30      \\
LUMO & 1.46     & \textbf{1.20}      \\
gap & 1.80     & \textbf{1.75}      \\
R2 & 10.87    & \textbf{2.41}      \\
ZPVE & 16.53    & \textbf{3.85}      \\
U0 & 3.16     & \textbf{0.92}      \\
U & 3.18     & \textbf{0.93}      \\
H & 3.20     & \textbf{0.93}      \\
G & 2.97     & \textbf{0.92}      \\
Cv & 2.17     & \textbf{1.28}      \\
Omega & 0.83     & \textbf{0.74}      \\
\hline
Average & 3.98 & \textbf{1.53} 
\end{tabular}
\end{table}

\begin{table}[t]
\centering
\caption{Performance With Different Input Information}
\label{tb:input}
\begin{tabular}{llll}
\hline
Target  & no distance & distance & dist + exp hydrogen \\
\hline
mu      & 3.81        & 0.95     & \textbf{0.30}                \\
alpha   & 2.04        & 1.18     & \textbf{0.92}                \\
HOMO    & 1.71        & 1.10     & \textbf{0.99}                \\
LUMO    & 1.41        & 1.06     & \textbf{0.87}                \\
gap     & 2.26        & 1.74     & \textbf{1.60}                \\
R2      & 13.67       & 0.57     & \textbf{0.15}                \\
ZPVE    & 3.02        & 2.57     & \textbf{1.27}                \\
U0      & 0.72        & 0.55     & \textbf{0.45}                \\
U       & 0.79        & 0.55     & \textbf{0.45}                \\
H       & 0.80        & 0.59     & \textbf{0.39}                \\
G       & 0.74        & 0.55     & \textbf{0.44}                \\
Cv      & 1.71        & 0.99     & \textbf{0.80}                \\
Omega   & 0.78        & 0.41     & \textbf{0.19}                \\
\hline
Average & 2.57        & 0.98     & \textbf{0.68}               
\end{tabular}
\end{table}

\end{document}